\documentclass[sigconf]{acmart}

\usepackage{multirow}
\usepackage{enumitem}

\AtBeginDocument{%
  }

\copyrightyear{2023}
\acmYear{2023}
\setcopyright{acmlicensed}
\acmConference[MM '23] {Proceedings of the 31st ACM International Conference on Multimedia}{October 29--November 3, 2023}{Ottawa, ON, Canada.}
\acmBooktitle{Proceedings of the 31st ACM International Conference on Multimedia (MM '23), October 29--November 3, 2023, Ottawa, ON, Canada}
\acmPrice{15.00}
\acmISBN{979-8-4007-0108-5/23/10}
\acmDOI{10.1145/3581783.3612285}

\begin{document}
\begin{sloppypar}

\title{Text-based Person Search without Parallel Image-Text Data}

\author{Yang Bai}
\affiliation{
  \institution{Soochow University}
  \city{Suzhou}
  \country{China}
}
\email{ybaibyougert@stu.suda.edu.cn}

\author{Jingyao Wang}
\affiliation{
  \institution{Soochow University}
  \city{Suzhou}
  \country{China}
}
\email{luoyetingqiu181@gmail.com}

\author{Min Cao}
\authornote{Corresponding author.}
\affiliation{
  \institution{Soochow University}
  \city{Suzhou}
  \country{China}
}
\email{mcao@suda.edu.cn}

\author{Chen Chen}
\affiliation{
  \institution{Institute of Automation, Chinese Academy of Sciences}
  \city{Beijing}
  \country{China}
}
\email{chen.chen@ia.ac.cn}

\author{Ziqiang Cao}
\affiliation{
  \institution{Soochow University}
  \city{Suzhou}
  \country{China}
}
\email{zqcao@suda.edu.cn}

\author{Liqiang Nie}
\affiliation{
  \institution{Harbin Institute of Technology}
  \city{Shenzhen}
  \country{China}
}
\email{nieliqiang@gmail.com}

\author{Min Zhang}
\affiliation{
  \institution{Soochow University}
  \city{Suzhou}
  \country{China}
}
\email{minzhang@suda.edu.cn}

\renewcommand{\shortauthors}{Yang Bai et al.}

\begin{abstract}
Text-based person search (TBPS) aims to retrieve the images of the target person from a large image gallery based on a given natural language description.
Existing methods are dominated by training models with parallel image-text pairs, which are very costly to collect.
In this paper, we make the first attempt to explore TBPS without parallel image-text data ($\mu$-TBPS), in which only non-parallel images and texts, or even image-only data, can be adopted.
Towards this end, we propose a two-stage framework, \textit{\textbf{g}eneration-\textbf{t}hen-\textbf{r}etrieval} (GTR), to first generate the corresponding pseudo text for each image and then perform the retrieval in a supervised manner.
In the generation stage, we propose a fine-grained image captioning strategy to obtain an enriched description of the person image, which firstly utilizes a set of instruction prompts to activate the off-the-shelf pretrained vision-language model to capture and generate fine-grained person attributes, and then converts the extracted attributes into a textual description via the finetuned large language model or the hand-crafted template.
In the retrieval stage, considering the noise interference of the generated texts for training model, we develop a confidence score-based training scheme by enabling more reliable texts to contribute more during the training.
Experimental results on multiple TBPS benchmarks (i.e., CUHK-PEDES, ICFG-PEDES and RSTPReid) show that the proposed GTR can achieve a promising performance without relying on parallel image-text data.
\end{abstract}




\maketitle

\section{Introduction}
Text-based person search (TBPS) aims to retrieve the images of the target person from a large image gallery by a given textual description, which has many potential applications in modern surveillance systems (e.g., searching for suspects, lost children or elderly individuals).
This task is closely related to person re-identification~\cite{ye2021deep, leng2019survey} and image-text retrieval~\cite{ijcai2022-759, 10119165, sun2023hierarchical, sun2022feature}, yet exhibits unique characteristics and challenges.
Compared to person re-identification with image query, TBPS provides a more user-friendly search with the open-form text query, correspondingly eliciting the challenge of cross-modal modeling due to the modality heterogeneity.
Compared to the general image-text retrieval, TBPS focuses on cross-modal retrieval specific for the person with more fine-grained details, tending to larger intra-class variance as well as smaller inter-class variance, which toughly bottlenecks the retrieval performance.

\begin{figure}
  \centering
  \setlength{\abovecaptionskip}{0.1cm}
  \setlength{\belowcaptionskip}{-0.2cm}
  \includegraphics[width=0.95\linewidth]{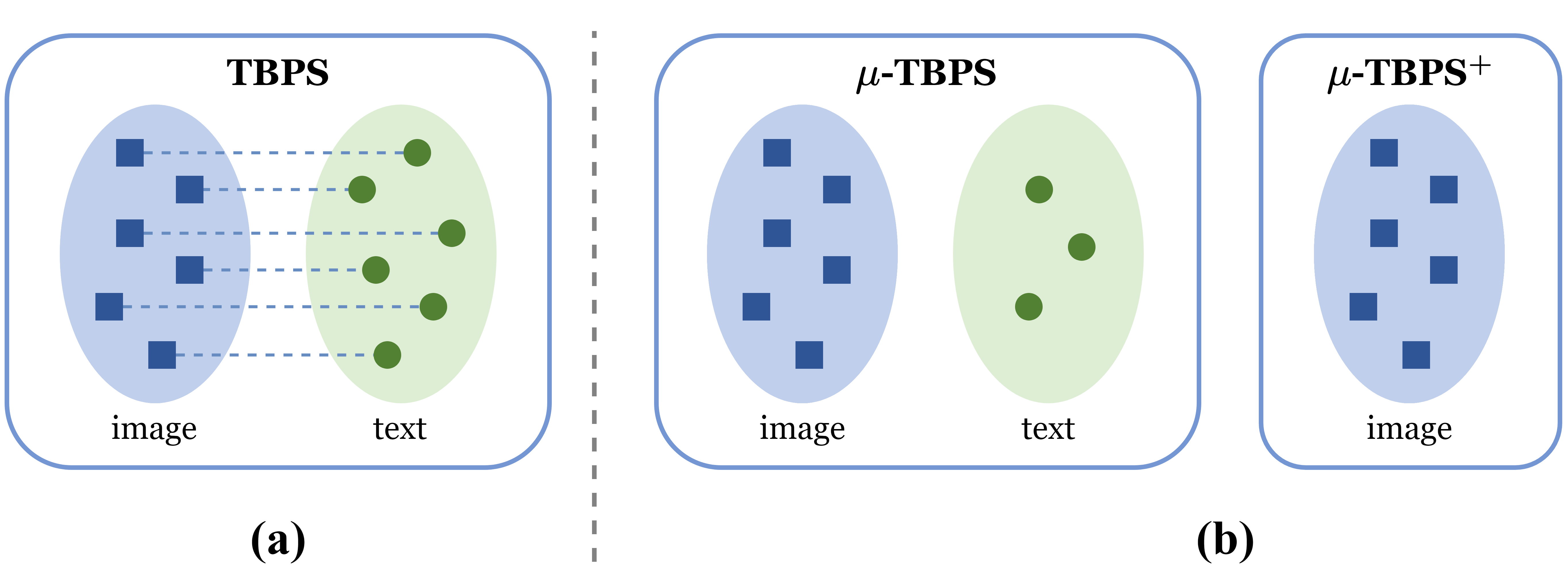}
  \caption{Illustration of 
            (a) canonical TBPS with parallel image-text pairs, 
            (b) TBPS without parallel image-text data. 
            Left: $\mu$-TBPS with separately collected non-parallel images and texts.
            Right: $\mu$-TBPS with image-only corpus ($\mu$-TBPS$^+$).}
  \label{fig1}
\end{figure}

To handle the aforementioned challenge, a series of TBPS works~\cite{li2017person, wang2020vitaa, shao2022learning, cvpr23crossmodal} are proposed to learn robust modality-invariant representations by designing sophisticated cross-modal alignment strategies.
These works heavily rely on the parallel image-text pairs for training the model, as illustrated in Figure~\ref{fig1} (a).
Although person images are relatively easily accessible via the deployed surveillance cameras, it is a labor-intensive and time-consuming process to annotate person images with textual descriptions.
In particular, \citeauthor{zhao2021weakly}~\cite{zhao2021weakly} propose the weakly supervised TBPS, which only releases the identity labeling for person images but still requires manually-labeled parallel image-text pairs for training.
\citeauthor{jing2020cross}~\cite{jing2020cross} propose the domain-adaptive TBPS, which has no dependency on the parallel image-text pair data in the target domain, while the data in the source domain still require human annotation.
This spontaneously raises a question: 
\textit{Can we well perform text-based person search without costly human annotation for parallel image-text data?}

Towards this end, we make the first attempt to explore text-based person search without parallel image-text data (called $\mu$-TBPS).
In this regime, the model is trained only based on the knowledge of the separately collected non-parallel images and texts about person\footnote{The two corpora are collected independently without correspondence: one consists of person images, and the other comprises the fine-grained textual descriptions with various attributes related to a person (e.g., appearance, wearing and accessories).}, as illustrated on the left-hand side of Figure~\ref{fig1} (b).
$\mu$-TBPS is more resource-saving due to no more reliance on human annotation, but also yields a significant challenge as the model is required to perform cross-modal alignment in the absence of cross-modal correspondence labels.
Furthermore, considering that person images are more easily collected than the texts via the surveillance cameras, we also explore the solution under $\mu$-TBPS with image-only corpus (called $\mu$-TBPS$^+$), as shown on the right-hand side of Figure~\ref{fig1} (b).
Compared to the primitive $\mu$-TBPS, the advanced $\mu$-TBPS$^+$ highlights a more practical scenario.

In response to the aforementioned regimes, this paper presents a two-stage framework, \textit{\textbf{g}eneration-\textbf{t}hen-\textbf{r}etrieval} (GTR), which firstly generates pseudo texts corresponding to the person images for remedying the absent annotation, and then trains a retrieval modal in a supervised manner.

\begin{figure}
  \centering
  \setlength{\belowcaptionskip}{-0.2cm}
  \includegraphics[width=0.95\linewidth]{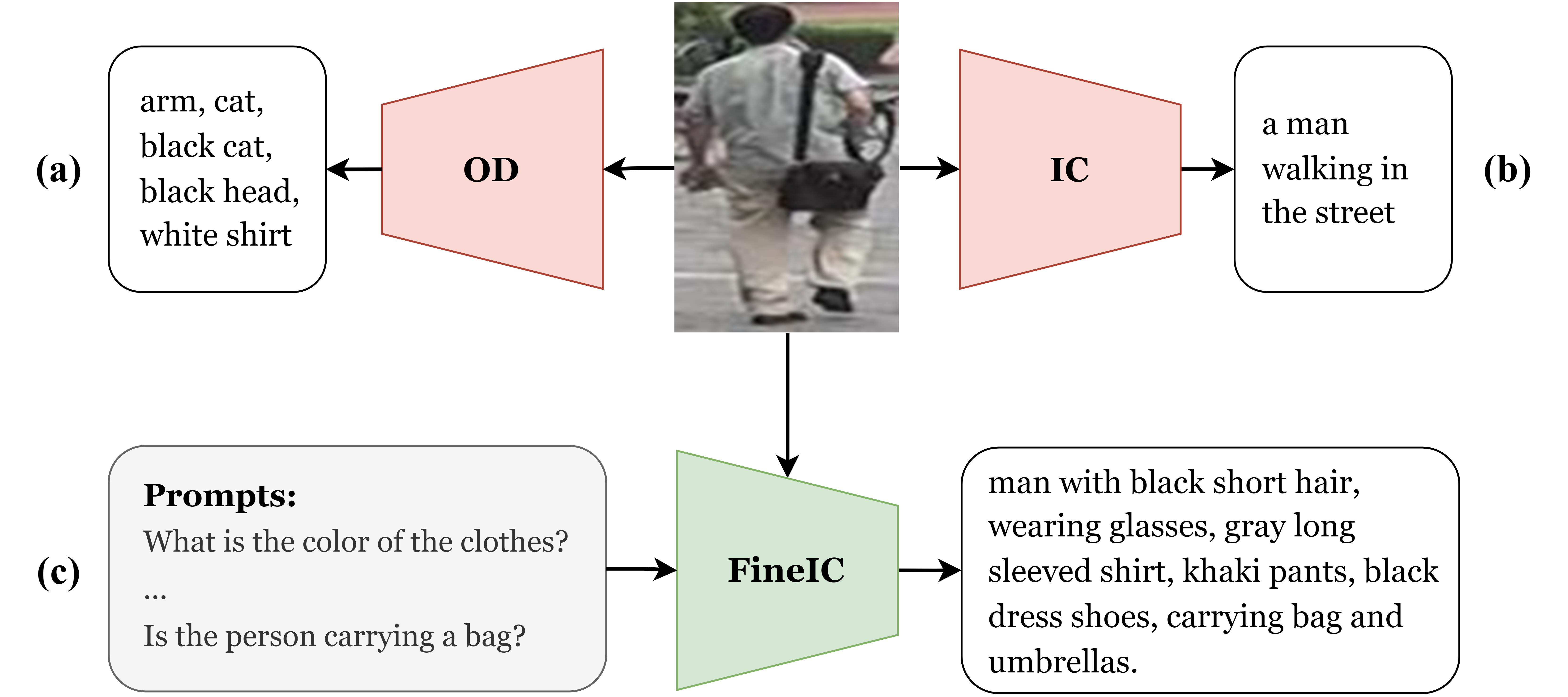}
  \caption{Illustration of text generation for a given person image.
           We illustrate (a) objection detection (OD) technology BUTD~\cite{anderson2018bottom} to extract the object tags and then take the tag sequence as the generated text,
           (b) the off-the-shelf vision-language model BLIP~\cite{li2022blip} to directly perform image captioning (IC),
           (c) the proposed fine-grained image captioning (FineIC) strategy to obtain an enriched description.}
  \Description{Three different technologies to perform text generation for a given person.}
  \label{fig2}
\end{figure}

\textbf{Generation}.
Given a person image, we aim to generate the corresponding pseudo textual description.
Referring to other cross-modal works without parallel image-text data~\cite{li2021unsupervised, zhou2022unsupervised, wang2022vlmixer, li2021unsupervised, dong2021unsupervised}, 
object detection technology is a straightforward and popular choice to perform the text generation.
Unfortunately, it is hard to obtain an expected result when applied to $\mu$-TBPS.
As depicted in Figure~\ref{fig2} (a), following the previous cross-modal works without parallel image-text data, we attempt to employ the widely-used object detector BUTD~\cite{anderson2018bottom} to extract the object tags from the image, and then take the tag sequence as the generated text.
However, the result is far from satisfactory.
For example, the tag ``arm'' is a common attribute for all individuals, which would yield no substantial contribution to TBPS due to the undifferentiated property.
Based on such tags, the generated pseudo text tends to become less distinctive with other texts due to the lack of informative and detailed descriptions.
Object detector excels at identifying general objects, which is naturally compatible with general cross-modal tasks but infeasible in this fine-grained TBPS task.
Alternatively, we can try the pretrained vision-language model to directly perform image captioning (IC). 
As shown in Figure~\ref{fig2} (b), the result is also unsuitable.
Since the vision-language model is pretrained on numerous image-text pairs with general content, the downstream image captioning is incapable of capturing the fine-grained details of a person.

To get an enriched description with fine-grained information, we propose a fine-grained image captioning (FineIC) strategy.
Firstly, considering that the attribute items about person appearance are typically finite (e.g., gender, type of clothes, color of clothes, existence of accessories), we design a set of corresponding instruction prompts to activate the vision-language model to capture the finite fine-grained person attributes, denoting image-to-attributes (I2A) extraction. 
For the gap between the attributes and the final open-form textual description, we fulfill the attributes-to-text (A2T) conversion by finetuning a language model (e.g., T5~\cite{raffel2020exploring}), during which the accessible text corpus is leveraged as a style reference to generate the final pseudo texts.
Furthermore, in the more practical scenario of $\mu$-TBPS$^+$ without accessible text corpus for reference, we design a hand-crafted template, in which we fill each blank with the proper attributes for the A2T conversion.

\textbf{Retrieval}.
In this stage, we use the images and the generated corresponding texts to train the retrieval model in a supervised manner. 
The retrieval model can, in principle, be adopted by any existing TBPS method.
Notably, the existing TBPS methods are trained with manually-annotated and well-aligned image-text pairs, while our constructed pairs are not always consistent since there inevitably exist incorrect depictions from the previous generation stage.
To alleviate the impact of the noise in training the retrieval modal, we develop a confidence score-based training (CS-Training) scheme. 
We collect the confidence score of the pseudo text in the generation stage, which is utilized to calibrate the propagation error in this stage.
Specifically, different weights are endowed to the pseudo texts based on the corresponding confidence score in the loss function of the retrieval model, enabling more confident texts to contribute more during training.
The proposed confidence score-based training scheme can be well established due to the flexible plug-and-play and the parameter-free characteristic.

Overall, the main contributions can be summarized as follows:
\begin{itemize}[leftmargin=20pt]
\item To the best of our knowledge, we make the first attempt to explore text-based person search without parallel image-text data.
We propose a two-stage framework GTR to first generate the pseudo texts for each image and then make the retrieval.
\item In the generation stage, we propose a fine-grained image captioning strategy to obtain the enriched textual descriptions, which first utilizes a set of instruction prompts to activate the vision-language models to generate fine-grained person attributes, and then adopts the finetuned language model or hand-crafted template to convert the attributes into fluent textual descriptions.
\item In the retrieval stage, considering the noise interference of the generated pseudo texts, we develop a confidence score-based training scheme by endowing more weights to more confident texts in the loss function of the retrieval model.
\item Experimental results on multiple TBPS benchmarks demonstrate that our method can achieve a promising performance without relying on parallel image-text data.
\end{itemize}

\section{Related work}
\subsection{Text-based Person Search}
Since \citeauthor{li2017person}~\cite{li2017person} introduce the TBPS task and publish the first public benchmark dataset CUHK-PEDES, recent years have witnessed its growing flourish.
Roughly, according to the focus, existing methods can be categorized into two streams: cross-modal alignment and representation learning.
The first group is dedicated to cross-modal alignment, which aims to align visual and textual features into a joint embedding space.
Earlier works depend on simple global alignment~\cite{zheng2020dual, farooq2020convolutional}, and then gradually evolve to multi-granularity correspondences~\cite{chen2018improvingdeep, chen2022tipcb, suo2022simple}.
Therein, several works~\cite{wang2020vitaa, jing2020pose, zheng2020hierarchical} resort to external technologies (e.g., human parsing, pose estimation and NLTK toolbox~\cite{loper2002nltk}) to extract fine-grained information.
Besides, a remarkable progress~\cite{li2022learning, gao2021contextual, gao2022conditional, ji2022asymmetric} has been made in self-adaptively semantic alignment.
The second group attempts to learn better modality-invariant representations.
\citeauthor{zhu2021dssl}~\cite{zhu2021dssl} propose to separate the person from the surroundings to reduce the misleading information.
Based on the observation that color plays a pivotal role in TBPS, \citeauthor{wu2021lapscore}~\cite{wu2021lapscore} develop two color-reasoning sub-tasks to explicitly build bidirectional fine-grained cross-modal associations.
\citeauthor{wang2022caibc}~\cite{wang2022caibc} notice the color over-reliance problem and propose a color deprivation and masking module to capture all-round information beyond color.
\citeauthor{shao2022learning}~\cite{shao2022learning} propose a granularity-unified representation learning framework to alleviate the granularity gap between two modalities.
More recently, a growing number of works~\cite{cvpr23crossmodal, han2021text, yan2022clip, wei2023calibrating} resort to large-scale vision-language pretrained models (e.g., CLIP~\cite{radford2021learning}) to inherit the general cross-modal knowledge to facilitate this fine-grained retrieval.
Therein, the most representative work IRRA~\cite{cvpr23crossmodal} additionally introduces a multi-modal interaction encoder upon CLIP~\cite{radford2021learning} to perform cross-modal implicit relation reasoning.
However, all of the aforementioned methods heavily rely on parallel image-text pairs, which require expensive and labor-intensive human annotation.

In fact, a handful of researchers have gradually noticed the burdensome annotating problem and begun to explore non-traditional TBPS.
\citeauthor{zhao2021weakly}~\cite{zhao2021weakly} pioneer weakly supervised TBPS without identity labeling and utilize the mutual refined clustering to generate pseudo labels, while the requirement of parallel image-text pairs still hampers its practical application. 
\citeauthor{jing2020cross}~\cite{jing2020cross} explore TBPS in a domain-adaptive setting to adapt the model to new target domains in the absence of parallel image-text data.
To this end, a moment alignment network is proposed to learn domain-invariant and modality-invariant representations.
While the domain-adaptive TBPS indeed does not require annotation in the target domain, it still relies on parallel image-text pairs in the source domain.
Different from the above works, in this paper, we make the first attempt to explore TBPS without relying on any parallel image-text data.

\begin{figure*}
  \centering
  \setlength{\belowcaptionskip}{-0.2cm}
  \includegraphics[width=0.9\linewidth]{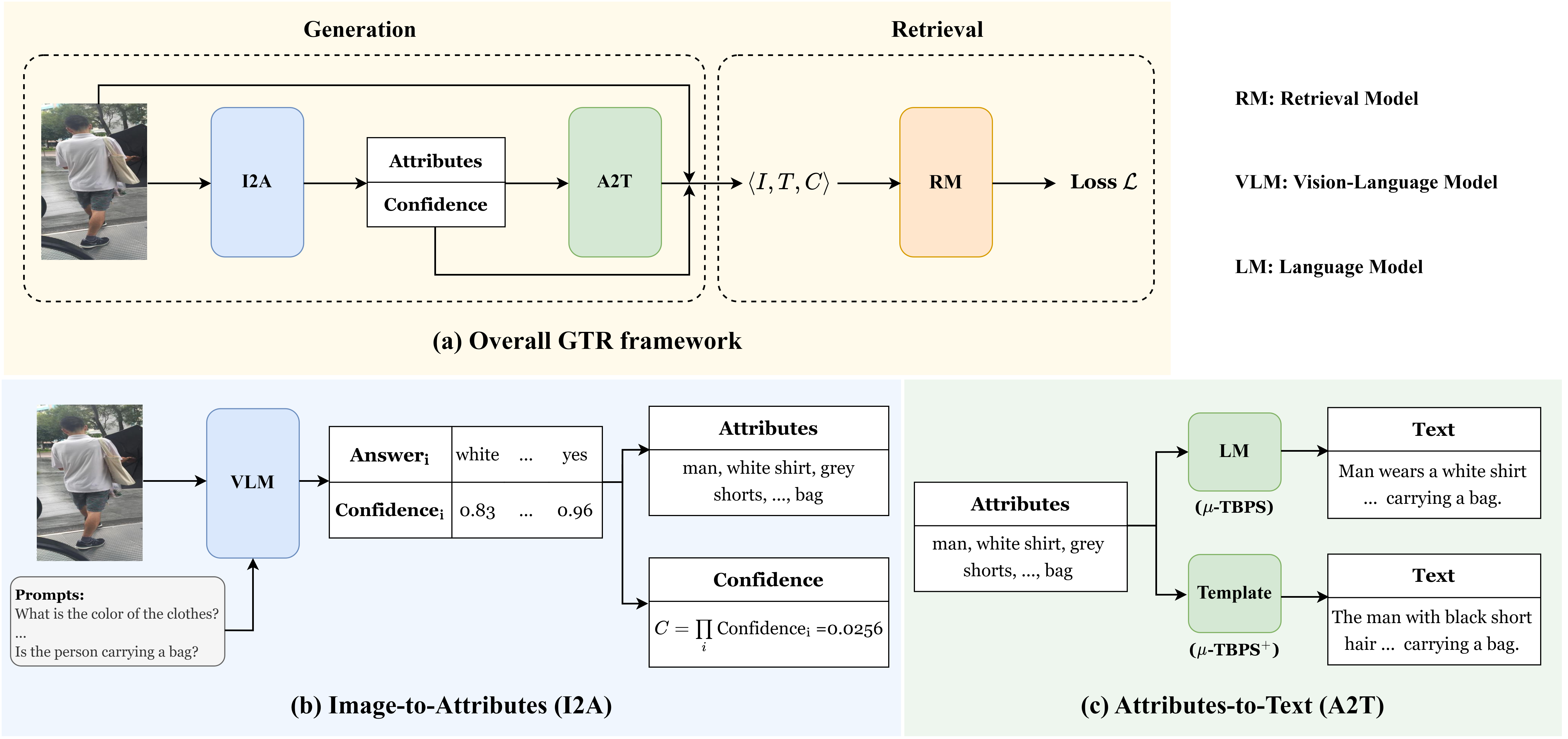}
  \caption{Overview of the proposed GTR framework (shown in (a)), which consists of two stages: generation and retrieval.
            In the generation stage, we propose a fine-grained image captioning (FineIC) strategy.
            Firstly, FineIC utilizes a set of instruction prompts to activate the vision language model to capture the fine-grained attributes for the image-to-attributes (I2A) extraction (shown in (b)).
            Then, the attributes-to-text (A2T) conversion (shown in (c)) is performed through the finetuned language model in $\mu$-TBPS or the hand-crafted template in $\mu$-TBPS$^+$.
            In the retrieval stage, we propose a confidence score-based training (CS-Training) scheme to train the retrieval model in a supervised manner.}
  \Description{Illustration of GTR framework.}
  \label{fig3}
\end{figure*}

\subsection{Unsupervised Vision-Language Tasks}
Considering that the parallel image-text data are extremely labor-intensive to collect, the persistent endeavor is dedicated to unsupervised learning in vision-language tasks without parallel image-text data, such as unsupervised vision-language pretraining~\cite{li2021unsupervised, zhou2022unsupervised, chen2022end, wang2022vlmixer}, unsupervised image captioning~\cite{feng2019unsupervised, gu2019unpaired, guo2021recurrent, ben2021unpaired, laina2019towards, liu2019exploring}, and unsupervised text-to-image synthesis~\cite{dong2021unsupervised}.
In unsupervised vision-language pretraining, U-VisualBERT~\cite{li2021unsupervised} makes the first exploration by conducting the masked prediction on image-only and text-only corpora, and introducing the object detection tags as anchor points to bridge the two modalities.
$\mu$-VLA~\cite{zhou2022unsupervised} adopts multi-granular alignment with a retrieved weakly aligned image-text corpus.
\citeauthor{wang2022vlmixer}~\cite{wang2022vlmixer} propose a novel data augmentation strategy, cross-modal CutMix (CMC), to transform natural sentences from the textual view to a multi-modal view through a pre-established image patch gallery, which is constructed from the object regions and their corresponding tags.
For unsupervised image captioning, \citeauthor{feng2019unsupervised}~\cite{feng2019unsupervised} make the first exploration and adopt a generative adversarial network, where the visual concepts from an object detector are used as the adversarial reward.
\citeauthor{guo2021recurrent}~\cite{guo2021recurrent} propose a memory-based method $R^2M$ to first convert words to a sentence in a supervised manner on text-only corpus, and then transfer the visual concepts extracted from an object detector to a language description in an unsupervised fashion on images.
For unsupervised text-to-image synthesis, \citeauthor{dong2021unsupervised}~\cite{dong2021unsupervised} make the first attempt and utilize the visual concepts to bridge two modalities: a sequence-to-sequence model is first trained to convert concept words to a sentence, and the visual concepts detected from an image are then directly inferred into a language description by the trained model.

We can clearly see that all the above works consistently rely on a critical technology: object detection.
However, when applied to this fine-grained task TBPS, object detection is hard to yield a satisfactory result.
The fundamental issue arises due to the significant disparity in tasks.
Therein, the object detector is trained on generic classes, inherently consistent with the above general vision-language tasks; while TBPS highlights more fine-grained categories such as various types of clothing, trousers, and footwear, leading to compatible issues with object detection.
In this paper, we explore an alternative viable solution and design a set of instruction prompts to activate the vision-language models to extract fine-grained attributes.

\section{Method}
Text-based person search without parallel image-text data ($\mu$-TBPS) relies on the separately collected images $I = \{I_1, I_2, \ldots, I_{N_i}\}$ and texts $S = \{S_1, S_2, \ldots, S_{N_s}\}$, or even image-only data $I$, where $N_i$ and $N_s$ are the total number of images and texts, respectively.
In the following, we formally delineate the proposed framework, generation-then-retrieval (GTR), for the two settings.
The overall framework is depicted in Figure~\ref{fig3}.
For simplicity, we omit the symbolic subscripts and use $I$ and $S$ to represent an image and a text, respectively.

\subsection{Generation}
Since there are no parallel image-text pairs available in $\mu$-TBPS, the first stage is aimed at generating the pseudo texts for each person image.
However, the widely-used object detection (OD) and image captioning (IC) are both sub-optimal to capture the fine-grained person attributes.
To get an enriched description, in this paper, we propose a fine-grained image captioning (FineIC) strategy.

Firstly, considering that each person image can be decoupled into a set of common attributes, we perform an image-to-attributes (I2A) conversion by designing a set of corresponding instruction prompts.
The prompts are composed of a series of questions pertaining to the person attributes, as listed in Appendix~\ref{instruction prompts}.

Formally, we denote the instruction prompts as $P = \{P_1, P_2, \ldots, P_{N_p}\}$, where $N_p$ is the number of the prompts. 
Given a person image $I$, we feed it with each prompt $P_i$ into an off-the-shelf vision-language model to perform the vision question answering (VQA), thus obtaining the result $\{\left<A_{i}, C_{i}\right>\}_{i=1}^{N_p}$, where $A_{i}$ is the answer to the prompt and $C_{i}$ is the corresponding confidence score that will be used in the retrieval stage. 
Then, the person attributes are logically obtained according to the answers.

For the gap between the attributes and the open-form natural language description, we use the accessible texts $S$ as a style reference to fulfill the attributes-to-text (A2T) conversion, in which we finetune a language model to incorporate the style of $S$ into the conversion.
Before finetuning, we first construct <attributes, text> of each accessible text. 
Specifically, given a text $S=\{s_1, s_2, \dots, s_n\}$ from the accessible text corpus, where $s_i$ is $i$-th word in the text, we extract the noun phrases as the attributes using constituency parsing technology~\cite{joshi2018extending}.
The attributes are then formed as an attribute sequence $W=\{w_1, w_2, \dots, w_m\}$, where $w_i$ is $i$-th word in the sequence.
With the constructed <$W$, $S$>, we can train the language model to output the text $S$ according to the sequence $W$ by maximizing the log-likelihood:
\begin{equation}
\setlength{\abovedisplayskip}{5pt}
\setlength{\belowdisplayskip}{5pt}
\begin{split}
    \max_\theta \sum_{i=1}^n \log p_\theta(s_i\mid w_1, w_2, \dots, w_m, s_1, s_2, \dots, s_{i-1}),
\end{split}
\label{equ1}
\end{equation}
where $\theta$ denotes the parameters of the language model.
During inference, we use the finetuned language model to convert the attributes extracted from the image to a fluent text.

Furthermore, for the more practical $\mu$-TBPS$^+$ without available texts, the above solution of finetuning the language model is not feasible. 
Therefore, we design a template for the A2T conversion:

\texttt{The $<$gender$>$ with $<$hair\_color$>$ $<$hair\_length$>$ hair wears $<$clothes\_color$>$ $<$clothes\_style$>$, $<$pants\_color$>$ $<$pants\_style$>$ and $<$shoes\_color$>$ $<$shoes\_style$>$. $<$He / She$>$ is carrying a $<$bag$>$, $<$glasses$>$, a $<$phone$>$ and an $<$umbrella$>$. The $<$gender$>$ is riding a $<$bike$>$.}


The first sentence in the template is designed for the fixed attributes, where the attribute blanks are filled by the corresponding extracted attributes.
In addition, the last two sentences are aimed at the variable attributes, in which we determine the presence of the particular words according to the extracted attributes.
The qualitative examples of the final constructed texts through the template are visualized in Appendix~\ref{visualization}.


To enrich the comprehensive information, we also generate a coupled description of the image via image captioning via the vision-language model.
We obtain the final pseudo text $T$ by concatenating the coupled description with the above generated text.

\subsection{Retrieval}
Through the previous generation stage, we obtain parallel image-text pairs that can be directly used to train the retrieval model in a supervised manner.
However, the pseudo text is not always well-aligned to the image, where the inevitable noise would predispose the retrieval model to the risk of misalignment.
To alleviate the impacts of the noise, we develop a confidence score-based training (CS-Training) scheme by introducing the confidence score of the pseudo text into the loss function of the retrieval model.

The confidence score of the generated text is a joint probability of the confidence of extracted attributes in the text (i.e., $C_i$ from the generation stage).
For simplicity, the attributes are deemed to be independent and identically distributed (i.i.d.). 
Thus, we obtain the confidence score of $T$ as:
\begin{equation}
    C = \prod_{i=1}^{N_p}C_{i}.
\label{confidence of text}
\end{equation}

The confidence score indirectly measures the consistency of the image-text pair $\left<I, T\right>$.
For this, we can adjust the contribution of each image-text pair by endowing the corresponding confidence score in the loss function.
Taking the retrieval model BLIP~\cite{li2022blip} as an example, the vanilla optimization objects include an image-text contrastive learning (ITC) loss and an image-text matching (ITM) loss.
Therein, ITC aims to pull the positive pairs together and push the negative pairs away, while ITM focuses on predicting whether an image-text pair is positive or negative.
We incorporate the confidence score into the ITC loss as:
\begin{equation}
\begin{split}
    \mathcal{L}_{itc}^{i2t} =& -\mathbb{E}_{(I, T, C)\sim D}C^\beta\log\frac{\exp{(s(I, T)/\tau)}}{\sum_{m=1}^M\exp{(s(I, T_m)/\tau)}},\\
    \mathcal{L}_{itc}^{t2i} =& -\mathbb{E}_{(I, T, C)\sim D}C^\beta\log\frac{\exp{(s(I, T)/\tau)}}{\sum_{m=1}^M\exp{(s(I_m, T)/\tau)}},\\
    &\qquad\mathcal{L}_{itc} = \left(\mathcal{L}_{itc}^{i2t} + \mathcal{L}_{itc}^{t2i}\right)\ /\ 2,
\end{split}
\label{cs-based ITC}
\end{equation}
where $M$ is the number of instances in a mini-batch, $s(I, T)$ measures the cosine similarity between the image $I$ and the text $T$, $\tau$ is a learnable temperature, and $\beta$ is a hyper-parameter to control the importance of the confidence score $C$.
The confidence score-based ITM loss is defined as:
\begin{equation}
    \mathcal{L}_{itm} = \mathbb{E}_{(I, T, C)\sim D}C^\beta{\mathcal{H}}\left(y, \phi(I, T)\right),
    \label{cs-based ITM}
\end{equation}
where $\mathcal{H}$ denotes the cross-entropy function. $y$ is $2$-dimension one-hot vector representing the ground-truth label (i.e., $\left[0, 1\right]$ for the positive pairs, and $\left[1, 0\right]$ for the negative pairs). $\phi(I, T)$ means the predicted matching probability of the pair.

With the integration of the confidence score, more confident image-text pair will offer more contribution during the training.
Notably, when $\beta$ is set to $0$, the confidence score-based loss will degenerate into the vanilla one, which means the confidence score is ignored.

\begin{table*}
\centering
\tabcolsep=10pt
\caption{GTR-Baseline and Ablation study on each component of the proposed GTR on CUHK-PEDES.}
\begin{tabular}{l|cccc|ccc|c}
\toprule
\multirow{2}{*}{\textbf{Methods}}  & \multicolumn{3}{c}{\textbf{Generation}} & \textbf{Retrieval} & \multirow{2}{*}{\textbf{R@1}} & \multirow{2}{*}{\textbf{R@5}} & \multirow{2}{*}{\textbf{R@10}} & \multirow{2}{*}{\textbf{mAP}} \\
                        & \textbf{IC} & \textbf{I2A} & \textbf{A2T} & \textbf{CS-Training} &  &  &  &  \\
\midrule
GTR-Baseline   &     & OD     &      &     & 31.03 & 49.81 & 59.21 & 27.65 \\
\midrule
M1             & \checkmark &            &            &            & 40.22 & 60.35 & 68.92 & 34.32 \\
M2             &            & \checkmark &            &            & 40.58 & 61.78 & 70.31 & 37.39 \\
M3             & \checkmark & \checkmark &            &            & 47.43 & 67.56 & 75.76 & 42.31 \\
M4 ($\mu$-TBPS) & \checkmark & \checkmark & \checkmark &           & 48.16 & 68.66 & 76.36 & 42.77 \\
M4 ($\mu$-TBPS$^+$) & \checkmark & \checkmark & \checkmark &       & 47.27 & 67.58 & 75.62 & 42.28 \\
\midrule
GTR ($\mu$-TBPS)   & \checkmark & \checkmark & \checkmark & \checkmark & 48.49 & 68.88 & 76.51 & 43.67 \\
GTR ($\mu$-TBPS$^+$) & \checkmark & \checkmark & \checkmark & \checkmark & 47.53 & 68.23 & 75.91 & 42.91 \\
\bottomrule
\end{tabular}
\label{table1}
\end{table*}

\section{Experiments}

\subsection{Datasets}
\textbf{CUHK-PEDES}~\cite{li2017person} is the most commonly-used dataset in TBPS. It consists of $40,206$ images and $80,440$ texts from $13,003$ identities in total, which are split into $34,054$ images and $68,126$ texts from $11,003$ identities in the training set, $3,078$ images and $6,158$ texts from $1,000$ identities in the validation set, and $3,074$ images and $6,156$ texts from $1,000$ identities in the test set. The average length of all texts is $23$. 
\noindent\textbf{ICFG-PEDES}~\cite{ding2021semantically} contains $54,522$ images from $4,102$ identities in total. Each of the images is described by one text. The dataset is split into $34,674$ images from $3,102$ identities in the training set, and $19,848$ images from $1,000$ identities in the test set. On average, there are $37$ words for each text.
\noindent\textbf{RSTPReid}~\cite{zhu2021dssl} consists of $20,505$ images of $4,101$ identities. Each identity has $5$ corresponding images captured from different cameras. Each image is annotated with $2$ textual descriptions, and each description is no shorter than $23$ words. There are $3,701$/$200$/$200$ identities utilized for training/validation/testing, respectively.

During the experiments, we evaluate the proposed method on each dataset without using the parallel image-text data.
Instead, we employ its training images as the image corpus, and the captions of the training sets from the three datasets without the cross-modal correspondence information as the accessible text corpus in $\mu$-TBPS.
Certainly, we only take the image corpus to perform $\mu$-TBPS$^+$.

\subsection{Protocol}
We adopt the widely-used Rank@K (R@K for short, K=$1,5,10$) metric to evaluate the performance of the proposed method. 
Specifically, given a query text, we rank all the test images via the similarity with the query text, and the search is deemed to be successful if top-K images contain any corresponding identity. 
R@K is the percentage of successful searches. 
In addition, we also adopt the mean average precision (mAP) as a complementary metric.
Rank@K reflects the accuracy of the first few retrieval results, while mAP emphasizes the comprehensive performance of the method.

\subsection{Implementation Details}
We conduct all experiments on $4$ NVIDIA A$40$ GPUs.
In the generation stage, the proposed FineIC utilizes $14$ instruction prompts (as detailed in Appendix~\ref{instruction prompts}) to activate the vision-language model for the I2A extraction, and then uses the finetuned language model or hand-crafted template for the A2T conversion.
In the retrieval stage, we use the constructed parallel image-text pairs to train the retrieval model under the proposed CS-Training scheme.
Generically, the models in each stage can be substituted by existing methods, which are experimentally verified in Section~\ref{Generalization Ability}.

For the generation stage, we use BLIP~\cite{li2022blip} and T5~\cite{raffel2020exploring} as the default models in I2A and A2T, respectively.
BLIP is a vision-language model that adopts a multi-modal mixture of encoder-decoder via the weight-sharing mechanism.
It bootstraps the generated synthetic captions while removing the noisy ones, eventually achieving a remarkable performance in various cross-modal tasks.
T5 is a language model. It adopts a Transformer~\cite{vaswani2017attention} encoder-decoder architecture and is proposed as a unified model that converts all text-based language problems into a text-to-text format by designing the specific task prompts.
We finetune T5 with a learning rate of $3e\!-\!5$ and a batch size of $20$ for $50$ epochs.
For the retrieval stage, we use BLIP as the default retrieval model, which is trained with a batch size of $52$ for $30$ epochs.
We use random horizontal flipping as the data augmentation.
The hyper-parameter $\beta$ in Equation~\eqref{cs-based ITC} and \eqref{cs-based ITM} is set as $0.8$ to control the importance of the confidence score.

\subsection{GTR-Baseline and Ablation Study}
In this section, we conduct a series of ablation experiments on CUHK-PEDES~\cite{li2017person} to analyze the superiority and effectiveness of the proposed GTR framework.
Since text-based person search without parallel image-text data is a completely novel task setting and has never been explored before, we first construct a baseline under the same settings as GTR for a fair comparison.

\subsubsection{GTR-Baseline}
\hfill \\
We construct a baseline that also employs the generation-then-retrieval pipeline, where the main difference with the proposed GTR lies in the generation solution.
Following the vision-language works without parallel image-text data~\cite{chen2022end, li2021unsupervised}, we utilize the widely-used object detector BUTD~\cite{anderson2018bottom} to extract the object tags of the image, and use the tag sequence as the generated pseudo text.
Therein, we set the minimum number of detected tags as $5$, the maximum number as $15$, and the attribute confidence threshold as $0.6$.

\subsubsection{Effectiveness of GTR Components}
\hfill \\
In this part, we verify the contribution of each component of GTR, as shown in Table~\ref{table1}.
The constructed GTR-Baseline relies on the object detector BUTD~\cite{anderson2018bottom} to extract person attributes for generating the pseudo texts.
The experimental results quantitatively demonstrate the inferiority of object detection responding to this particular fine-grained retrieval task.
When only performing IC, the generated texts focus on the comprehensive description of the person image.
Without the fine-grained depiction, IC only attains a modest performance of $34.32\%$ at mAP.
Hence, we propose the FineIC strategy to get more enriched descriptions, which consists of I2A extraction and A2T conversion.
To evaluate the effectiveness of I2A, we use the extracted attributes to form a text sequence as the generated text, which brings a considerable performance of $37.39\%$ at mAP.
When incorporating the comprehensive information from IC, the performance has a remarkable advance (e.g., from $40.58\%$ to $47.43\%$ at R@1).
Then, in order to narrow the gap between the attributes and the open-form textual descriptions, the proposed FineIC adopts the finetuned language model to fulfill the A2T conversion in $\mu$-TBPS, which eventually achieves a significant improvement (e.g., from $47.43\%$ to $48.16\%$ at R@1).
For the more practical $\mu$-TBPS$^+$, the well-crafted template is designed for the A2T conversion.
The results show that this hand-crafted template seems to have little discernible effect.
We conjecture that the designed template is too rigid to bring additional information gain to the retrieval.
Furthermore, taking into consideration that the generated pseudo texts are not always well-aligned to the images, we propose the CS-Training scheme by adjusting the weights of the texts in the loss function.
The experimental results clearly demonstrate the effectiveness of the CS-Training scheme (e.g., a considerable advance of $0.90\%$ in $\mu$-TBPS and $0.63\%$ in $\mu$-TBPS$^+$ at mAP).
With the synergy of each component, the proposed GTR eventually achieves a promising performance of $48.49\%$ in $\mu$-TBPS and $47.53\%$ in $\mu$-TBPS$^+$ at R@1.

\subsubsection{Generalization Ability}
\label{Generalization Ability}
\hfill \\
We propose a well-adapted framework for $\mu$-TBPS, in which the models can be compatibly adopted by existing methods.
To verify the generation ability of the proposed framework, we conduct a series of experiments with different variants, including OFA~\cite{wang2022ofa} for I2A, GPT-2~\cite{radford2019language} for A2T, ALBEF~\cite{li2021align} and IRRA~\cite{cvpr23crossmodal} for the retrieval.
The experimental results are shown in Table~\ref{table2}.
\begin{itemize}[leftmargin=*, nolistsep]
    \item \textbf{I2A}.
    OFA~\cite{wang2022ofa} is a ``One For All'' vision-language model, which unifies a diverse set of cross-modal and unimodal tasks via a sequence-to-sequence learning framework with a unified instruction-based task representation.
    The proposed GTR with OFA performs roughly on par with that with the default BLIP.
    Specifically, BLIP exhibits a superior I2A performance over OFA, leading to preferable attribute extraction and yielding slightly better retrieval results.
    The results also indicate that the proposed GTR has the potential to achieve an even higher level with the emergence of a more effective vision-language model.

    \item \textbf{A2T}.
    Similar to the default T5~\cite{raffel2020exploring}, GPT-2~\cite{radford2019language} is also a well-known and highly successful pretrained language model in the field of natural language processing.
    It employs a Transformer~\cite{vaswani2017attention} decoder architecture rather than the encoder-decoder structure in T5.
    The different architectures give rise to their distinct specializations: GPT-2 has a knack for continuous text generation (e.g., articles, stories and conversations), while T5 is proficient in input-to-output mapping (e.g., language translation and text summarization).
    In the specific application scenario of A2T conversion, which can be deemed as attributes-to-text mapping, T5 obviously could achieve a better performance than GPT-2, in line with the results in Table~\ref{table2}.

    \item \textbf{Retrieval}.
    ALBEF~\cite{li2021align} is a vision-language model which utilizes a contrastive loss to align the visual and textual representations before the deep fusion based on cross-modal attention. 
    IRRA~\cite{cvpr23crossmodal} is a retrieval model specific for TBPS.
    It also adopts an implicit relation reasoning module to enhance the fine-grained interaction through cross-modal attention, where a similarity distribution matching is proposed to enlarge the correlation between the matched pairs before the interaction.
    The modified confidence score-based losses of ALBEF and IRRA are shown in Appendix~\ref{confidence score-based loss}.
    It is clear from Table~\ref{table2} that the two models both exhibit promising retrieval results without reliance on parallel image-text pairs.
    Therein, IRRA achieves a relatively mediocre performance.
    We conjecture that this may be attributed to its greater susceptibility to misalignment and overfitting caused by the inevitable noise from the generated texts.
\end{itemize}

\begin{table}
\small
\centering
\tabcolsep=3.7pt
\caption{Ablations of generalization ability with different variants. Tmpl means the hand-craft template used in $\mu$-TBPS$^+$. The default settings are in bold.}
\begin{tabular}{c|cc|c|ccc|c}
\toprule
\multirow{2}{*}{\textbf{Mode}} & \multicolumn{2}{c|}{\textbf{Generation}} & \multirow{2}{*}{\textbf{Retrieval}} & \multirow{2}{*}{\textbf{R@1}} & \multirow{2}{*}{\textbf{R@5}} & \multirow{2}{*}{\textbf{R@10}} & \multirow{2}{*}{\textbf{mAP}} \\
& \textbf{I2A}            & \textbf{A2T}           &     &    &    &    &   \\
\midrule
\multirow{5}{*}{$\mu$-TBPS}  & \textbf{BLIP}   & \textbf{T5}    & \textbf{BLIP}  & \textbf{48.49}  & \textbf{68.88}   & \textbf{76.51}   &  \textbf{43.67} \\
                                & OFA    & T5      & BLIP    & 47.27   & 66.95    & 75.00    & 42.36     \\
                                & BLIP   & GPT-2   & BLIP    & 44.87   & 64.54    & 72.95    & 40.18    \\
                                & BLIP   & T5      & ALBEF    & 42.02   & 62.61    & 70.68    & 36.92     \\
                                & BLIP   & T5      & IRRA     & 30.72   & 51.20    & 61.42    & 28.24     \\
\midrule
\multirow{4}{*}{$\mu$-TBPS$^+$}  & \textbf{BLIP}   & \textbf{Tmpl}  & \textbf{BLIP}  & \textbf{47.53}  & \textbf{68.23}   & \textbf{75.91}   &  \textbf{42.91} \\
                                & OFA    & Tmpl    & BLIP    & 47.11   & 67.32    & 75.00    & 42.27     \\
                                & BLIP   & Tmpl    & ALBEF    & 40.69   & 61.18    & 69.69    & 37.53     \\
                                & BLIP   & Tmpl    & IRRA     & 30.26  & 51.27    & 61.31    & 28.15     \\
\bottomrule
\end{tabular}
\label{table2}
\end{table}

\begin{figure}
  \centering
  \includegraphics[width=\linewidth]{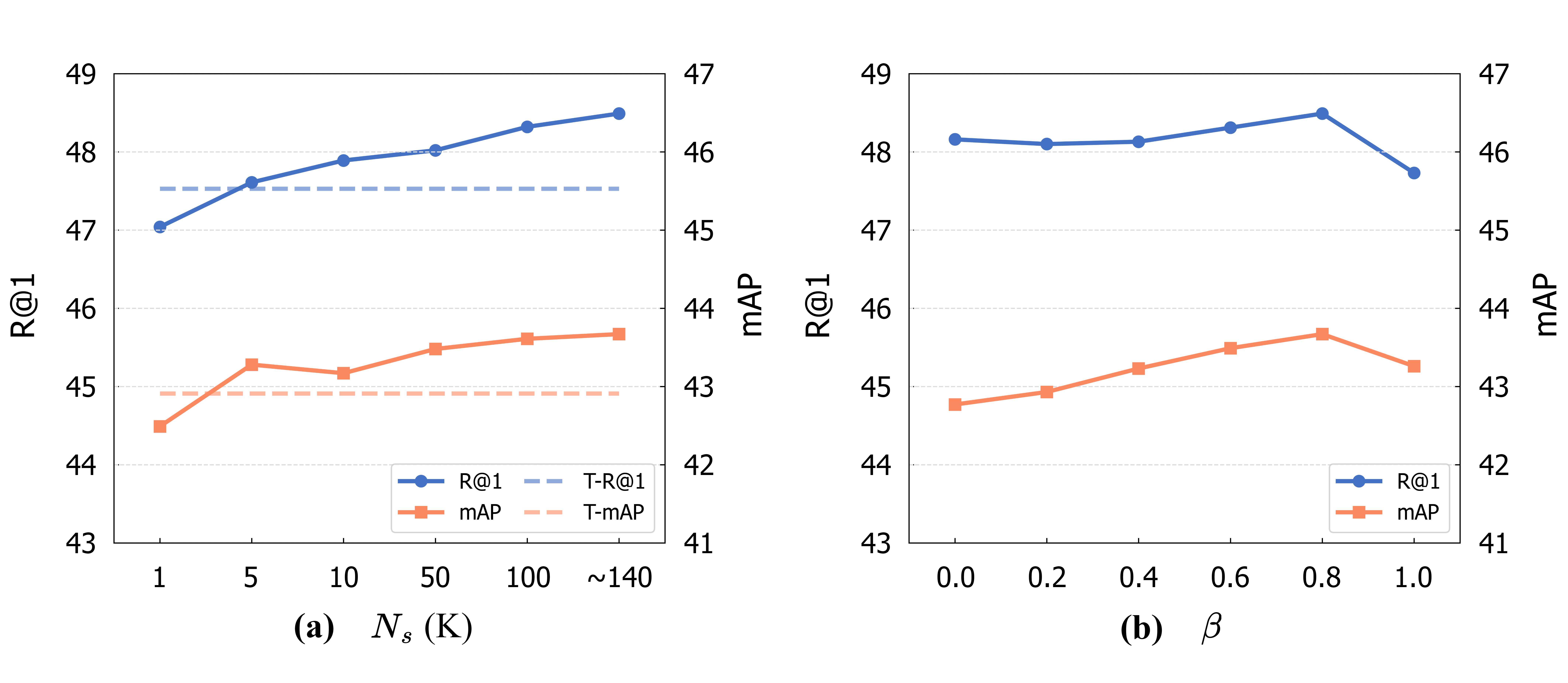}
  \caption{Influence of (a) the amount of the accessible texts $N_s$ in $\mu$-TBPS, where T-R@1 and T-mAP denote the performance of R@1 and mAP using the hand-crafted template, respectively. (b) the hyper-parameter $\beta$ to control the importance of the confidence score in the retrieval stage.}
  \label{fig4}
\end{figure}

\subsubsection{Amount of the Accessible Texts}
\hfill \\
When performing $\mu$-TBPS, the pseudo texts corresponding to the person images are generated by adopting the accessible texts from external sources as the style reference of the text. 
By default, the accessible texts are collected from the combined training texts in the three TBPS datasets in the experiment, and about $140$K of them. 
In this part, we investigate the influence of the amount of the accessible texts $N_s$ on the retrieval performance by taking a random sample of about $140$K texts.
From Figure~\ref{fig4} (a), we can see that the result presents a generally increasing trend as $N_s$ rises.
When $N_s$ exceeds about $5$K, the results have the advantage over text generation without the text reference (i.e., using the hand-crafted template for text generation).
It indicates that the introduction of the text reference can bring a competitive advantage on performance depending on its amount.
Small text references are hard to support the well training of a large language model for text generation.
Thereby, the results are inferior to ones without the text reference and using the hand-crafted template.
Moreover, compared to using the entire corpus (i.e., roughly $140$K texts) to achieve a higher performance of $48.49$ at R@1 and $43.67$ at mAP, we can adopt a cost-effective approach by collecting only $5$K sentences to attain a comparable level of $47.61$ at R@1 and $43.28$ at mAP.


\subsubsection{Hyper-parameter Analysis}
\hfill \\
We introduce a confidence score-based training scheme into the retrieval stage, where the hyper-parameter $\beta$ is used to control the importance of the confidence score.
The influence of $\beta$ is shown in Figure~\ref{fig4} (b).
Notably, $\beta=0$ means the confidence score is ignored, which attains an inferior performance of $48.16\%$ at R@1 and $42.77\%$ at mAP.
As $\beta$ increases, mAP presents an overall trend of initially rising and subsequently declining, and reaches the peak of $43.67\%$ at $\beta=0.8$.
For R@1, the performance initially exhibits a slight decline, followed by a gradual increase up to a peak of $48.49\%$ when $\beta$ is set to $0.8$, and eventually experiences a significant decrease.

\begin{table}
\small
\centering
\tabcolsep=5pt
\caption{Comparison with SOTA methods on CUHK-PEDES.}
\begin{tabular}{l|l|ccc|c}
\toprule
\textbf{Methods}      & \textbf{Reference}         & \textbf{R@1}   & \textbf{R@5}   & \textbf{R@10}  & \textbf{mAP}   \\
\midrule
\multicolumn{6}{c}{\textit{TBPS (w/ Parallel Image-Text Data)}} \\
\midrule
CNN-RNN~\cite{reed2016learning}      & CVPR-2016   & 8.07  & -     & 32.47 & -     \\
Neural Talk~\cite{vinyals2015show}  & CVPR-2015   & 13.66 & -     & 41.72 & -     \\
GNA-RNN~\cite{li2017person}      & CVPR-2017   & 19.05 & -     & 53.64 & -     \\
IATVM~\cite{li2017identity}        & ICCV-2017   & 25.94 & -     & 60.48 & -     \\
Dual Path~\cite{zheng2020dual}    & TOMM-2020   & 44.40 & 66.26 & 75.07 & -     \\
CMPM/C~\cite{zhang2018deep}       & ECCV-2018   & 49.37 & -     & 79.27 & -     \\
ViTAA~\cite{wang2020vitaa}        & ECCV-2020   & 55.97 & 75.84 & 83.52 & 51.60  \\
HGAN~\cite{zheng2020hierarchical}         & MM-2020     & 59.00 & 79.49 & 86.60 & -     \\
NAFS~\cite{gao2021contextual}         & arXiv-2021  & 59.94 & 79.86 & 86.70 & 54.07 \\
DSSL~\cite{zhu2021dssl}         & MM-2021     & 59.98 & 80.41 & 87.56 & -     \\
SSAN~\cite{ding2021semantically}         & arXiv-2021  & 61.37 & 80.15 & 86.73 & -     \\
LBUL~\cite{wang2022look}         & MM-2022     & 64.04 & 82.66 & 87.22 & -     \\
SAF~\cite{li2022learning}          & ICASSP-2022 & 64.13 & 82.62 & 88.40 & 58.61 \\
CAIBC~\cite{wang2022caibc}        & MM-2022     & 64.43 & 82.87 & 88.37 & -     \\
TIPCB~\cite{chen2022tipcb}        & Neuro-2022     & 64.26 & 83.19 & 89.10 & -     \\
LGUR~\cite{shao2022learning}         & MM-2022     & 65.25 & 83.12 & 89.00 & -     \\
BLIP~\cite{li2022blip}               & ICML-2022     & 65.61 & 82.84 & 88.65 & 58.02     \\
IRRA~\cite{cvpr23crossmodal}         & CVPR-2023   & 73.38 & 89.93 & 93.71 & 66.13 \\
\midrule
\multicolumn{6}{c}{\textit{$\mu$-TBPS (w/o Parallel Image-Text Data)}} \\
\midrule
GTR ($\mu$-TBPS)                        & \multicolumn{1}{c|}{-}   & 48.49 & 68.88 & 76.51 & 43.67 \\
GTR ($\mu$-TBPS$^+$)                    & \multicolumn{1}{c|}{-}   & 47.53 & 68.23 & 75.91 & 42.91 \\
\bottomrule
\end{tabular}
\label{table3}
\end{table}

\subsection{Comparison with SOTA Methods with Parallel Data}
We compare the proposed framework GTR to SOTA methods with parallel image-text data on CUHK-PEDES~\cite{li2017person}, ICFG-PEDES~\cite{ding2021semantically} and RSTPReid~\cite{zhu2021dssl}, as shown in Table~\ref{table3}, Table~\ref{table4} and Table~\ref{table5}, respectively.
Specifically, we report the results without re-ranking, in keeping with the proposed GTR (also without re-ranking).

Compared to these methods, the proposed GTR exhibits a performance gap, but also shows promising results.
The mainstream TBPS methods have achieved a significant advance for the past few years (e.g., from $8.07\%$ to $73.38\%$ at R@1 on CUHK-PEDES). 
Nonetheless, these methods heavily rely on human annotation to obtain parallel data for the training.
In contrast, the proposed GTR releases manual annotation and is trained without parallel image-text pairs.
We also notice that GTR has a relatively mediocre performance on ICFG-PEDES. 
The texts in ICFG-PEDES are much longer than those in other datasets, indicating a greater density of information in the text.
However, the generated texts are composed of limited attributes in GTR, which depend on the designed instruction prompts.
As a result, there is a degree of gap between the generated pseudo texts and the ground truth, eventually hampering the retrieval performance.

\section{Conclusion}
In this paper, we focus on performing text-based person search without costly human annotation
for parallel image-text data,
and make the first attempt to explore text-based person search without parallel image-text data.
For this, we propose a two-stage framework GTR to first remedy the absent annotation via generating corresponding pseudo texts and then train a retrieval modal in a supervised manner.
In the generation stage, we propose a fine-grained image captioning strategy to obtain the enriched description of the image, which firstly utilizes a set of instruction prompts to activate the vision-language model for extracting fine-grained person attributes, and then converts the attributes to a natural language description via the finetuned language model or the hand-crafted template.
In the retrieval stage, taking the inevitable noise of the generated texts into consideration, we develop a confidence score-based training scheme by endowing proper weights to the pairs of different consistency in the loss function of the retrieval model.
Eventually, the proposed GTR achieves a remarkable performance without the reliance on parallel image-text data on multiple TBPS benchmarks, which clearly demonstrates the superiority and effectiveness of the proposed method.

\begin{table}
\small
\centering
\tabcolsep=5pt
\caption{Comparison with SOTA methods on ICFG-PEDES.}
\begin{tabular}{l|l|ccc|c}
\toprule
\textbf{Methods}      & \textbf{Reference}         & \textbf{R@1}   & \textbf{R@5}   & \textbf{R@10}  & \textbf{mAP}   \\
\midrule
\multicolumn{6}{c}{\textit{TBPS (w/ Parallel Image-Text Data)}} \\
\midrule
BLIP~\cite{li2022blip}              & ICML-2022     &37.09      &55.19      &63.65      &21.39     \\
Dual Path~\cite{zheng2020dual}      & TOMM-2020     &38.99      &59.44      &68.41      &-      \\
CMPM/C~\cite{zhang2018deep}         & ECCV-2018     &43.51      &65.44      &74.26      &-      \\
ViTAA~\cite{wang2020vitaa}          & ECCV-2020     &50.98      &68.79      &75.78      &-      \\
TIPCB~\cite{chen2022tipcb}          & Neuro-2022    &54.96      &74.72      &81.89      &-      \\
SRCF~\cite{suo2022simple}           & ECCV-2022     &57.18      &75.01      &81.49      &-      \\
LGUR~\cite{shao2022learning}        & MM-2022       &57.42      &74.97      &81.45      &-      \\
IRRA~\cite{cvpr23crossmodal}        & CVPR-2023     &63.46      &80.25      &85.82      &38.06  \\
\midrule
\multicolumn{6}{c}{\textit{$\mu$-TBPS (w/o Parallel Image-Text Data)}} \\
\midrule
GTR ($\mu$-TBPS)                   & \multicolumn{1}{c|}{-}   & 29.64 & 47.23 & 55.54 & 14.20 \\
GTR ($\mu$-TBPS$^+$)                   & \multicolumn{1}{c|}{-}   & 28.25 & 45.21 & 53.51 & 13.82 \\
\bottomrule
\end{tabular}
\label{table4}
\end{table}

\begin{table}
\small
\centering
\tabcolsep=5pt
\caption{Comparison with SOTA methods on RSTPReid.}
\begin{tabular}{l|l|ccc|c}
\toprule
\textbf{Methods}      & \textbf{Reference}         & \textbf{R@1}   & \textbf{R@5}   & \textbf{R@10}  & \textbf{mAP}   \\
\midrule
\multicolumn{6}{c}{\textit{TBPS (w/ Parallel Image-Text Data)}} \\
\midrule
DSSL~\cite{zhu2021dssl}             & MM-2021     &39.05      &62.60      &73.95      &-      \\
SSAN~\cite{ding2021semantically}    & arXiv-2021  &43.50      &67.80      &77.15      &-      \\
LBUL~\cite{wang2022look}            & MM-2022     &45.55      &68.20      &77.85      &-      \\
IVT~\cite{shu2023see}               & ECCVW-2022  &46.70      &70.00      &78.80      &-      \\
BLIP~\cite{li2022blip}              & ICML-2022   &58.25      &77.85      &85.65      &44.08     \\
IRRA~\cite{cvpr23crossmodal}        & CVPR-2023   &60.20      &81.30      &88.20      &47.17  \\
\midrule
\multicolumn{6}{c}{\textit{$\mu$-TBPS (w/o Parallel Image-Text Data)}} \\
\midrule
GTR ($\mu$-TBPS)                   & \multicolumn{1}{c|}{-}   & 46.65 & 70.70 & 80.65 & 34.95 \\
GTR ($\mu$-TBPS$^+$)                   & \multicolumn{1}{c|}{-}   & 45.60 & 70.35 & 79.95 & 33.30 \\
\bottomrule
\end{tabular}
\label{table5}
\end{table}

\section{Broader Impact}
Text-based person search has many potential applications in the fields of intelligent surveillance, safety protection, smart city, etc.
However, existing methods heavily rely on parallel image-text pairs, which require costly and time-consuming human annotation to label the person images with natural language descriptions.
In this work, we make the first attempt to explore text-based person search without parallel image-text data.
We hope our work could effectively promote the development of text-based person search.
The potential negative impact lies in that the public datasets for text-based person search comprise surveillance images without formal consent, which may cause an invasion of privacy.
Hence, for both the data collection and utilization, further community endeavor is required to circumvent the negative impacts including but not limited to social bias, individual privacy and potential misuse.

\begin{acks}
This work is supported by the National Science Foundation of China under Grant NSFC 62002252, the National Science Foundation of China under Grant NSFC 62106165, and the foundation of Key Laboratory of Artificial Intelligence, Ministry of Education, P.R. China.
\end{acks}

\bibliographystyle{ACM-Reference-Format}
\balance
\bibliography{GTR}

\appendix

\section{Appendix}

\subsection{Instruction Prompts}
\label{instruction prompts}
To activate the large vision-language model to capture specific fine-grained details, we design $14$ instruction prompts pertaining to the person attributes as follows:

\begin{enumerate}
    \item[1.]   \texttt{What is the color of the clothes?}
    \item[2.]   \texttt{What is the style of the clothes?}
    \item[3.]   \texttt{What is the color of the pants?}
    \item[4.]   \texttt{What is the style of the pants?}
    \item[5.]   \texttt{What is the color of the shoes?}
    \item[6.]   \texttt{What is the style of the shoes?}
    \item[7.]   \texttt{What is the gender of the person?}
    \item[8.]   \texttt{What is the color of the hair?}
    \item[9.]   \texttt{Is the person with long hair?}
    \item[10.]  \texttt{Is the person wearing glasses?}
    \item[11.]  \texttt{Is the person holding a mobile phone?}
    \item[12.]  \texttt{Is the person holding an umbrella?}
    \item[13.]  \texttt{Is the person riding a bike?}
    \item[14.]  \texttt{Is the person carrying a bag?}
\end{enumerate}

\begin{figure*}
  \centering
  \includegraphics[width=0.95\linewidth]{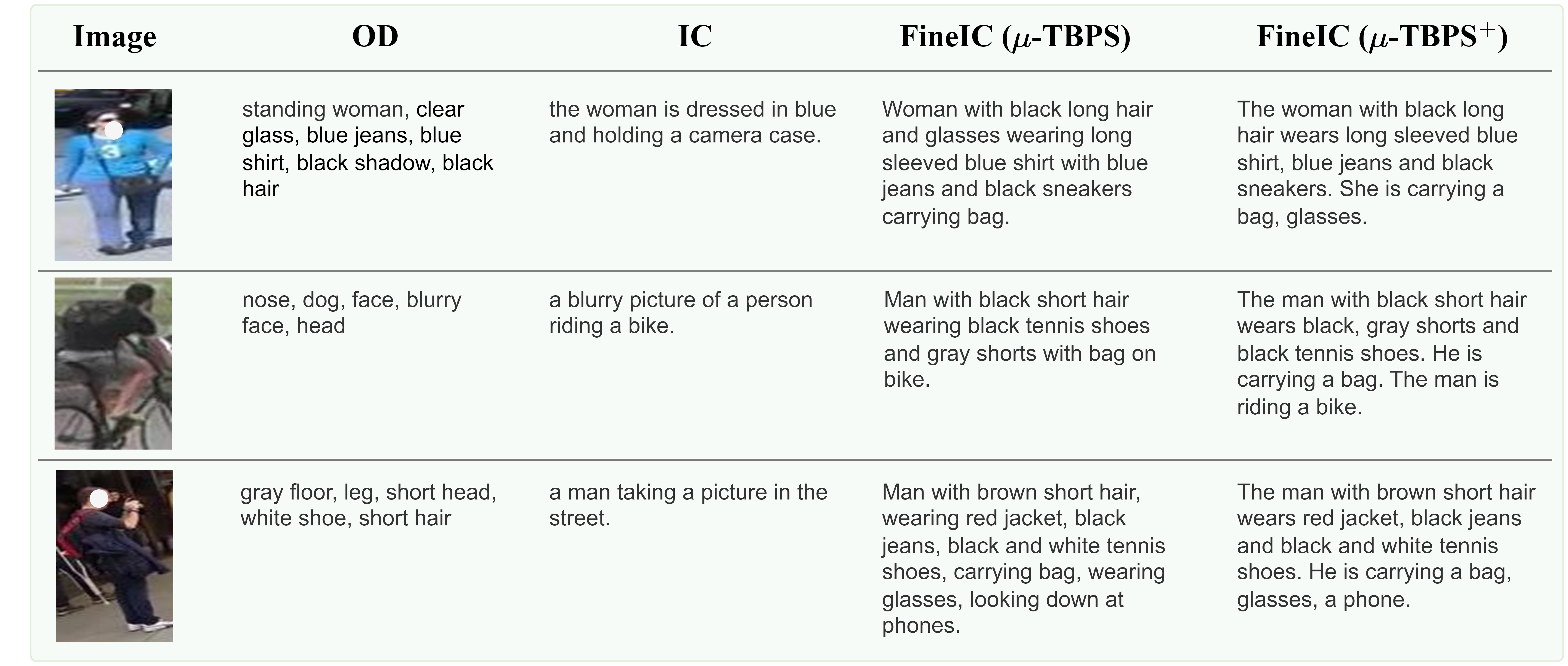}
  \caption{Visualization of generated texts from object detection (OD) by BUTD~\cite{anderson2018bottom}, image captioning (IC) on the vision-language model BLIP~\cite{li2022blip} and the proposed FineIC strategy.}
  \label{fig5}
\end{figure*}

\subsection{Confidence Score-based Loss}
\label{confidence score-based loss}
We conduct extensive experiments on more retrieval models to verify the generalization ability of the proposed GTR, including ALBEF~\cite{li2021align} and IRRA~\cite{cvpr23crossmodal}, as shown in Table~\ref{table2} of the paper.

\textbf{ALBEF}~\cite{li2021align} adopts the same loss functions as BLIP~\cite{li2022blip} for the retrieval task, i.e., image-text contrastive learning loss and image-text matching loss.
We incorporate the confidence score into the loss functions using the same approach as described in Equation~\eqref{cs-based ITC} and \eqref{cs-based ITM} of the paper.

\textbf{IRRA}~\cite{cvpr23crossmodal} employs three loss functions as the optimization objectives: implicit relation reasoning (IRR) loss, similarity distribution matching (SDM) loss and ID loss~\cite{zheng2020dual}.

IRR aims to predict the masked textual tokens by the rest of the unmasked tokens and the visual tokens, where the sampled embeddings of the masked tokens serve as an anchor to align the visual and textual contextualized representations.
We inject the confidence score into the IRR loss as:
\begin{equation}
\small
    \mathcal{L}_{irr} = -\frac{1}{N} \sum_{t=1}^N
                        \frac{C_{_t}^\beta}{|\mathcal{M}_{_t}||\mathcal{V}|} \sum_{i=1}^{|\mathcal{M}_{_t}|}\sum_{j=1}^{|\mathcal{V}|}
                        y_{_{tij}} \log\frac{\exp{(p_{_{tij}})}}{\sum_{k=1}^{|\mathcal{V}|}\exp{(p_{_{tik}})}},
\end{equation}
where $N$ is the number of instances in a mini-batch, $\mathcal{M}_{_t}$ denotes the set of the masked text tokens of the $t$-th text in the mini-batch, |$\mathcal{V}$| is the size of vocabulary $\mathcal{V}$, $p_{_{ti}}$ is the predicted probability distribution of the $i$-th masked token in the $t$-th text, and $y_{_{ti}}$ is a one-hot vocabulary distribution where the ground-truth token has a probability of $1$. $C_{_t}$ denotes the confidence score of the $t$-th text, and $\beta$ is a hyper-parameter to control the importance of the confidence score.

SDM loss is proposed to minimize the Kullback-Leibler (KL) divergence between image-text similarity distributions and the normalized ground-truth matching distributions.
Taking image-to-text matching as an example, the confidence score-based SDM loss is denoted as:
\begin{equation}
\begin{split}
    &p_{i,j} = \frac{\exp{(C_j^\beta s(f_i^v, f_j^t)/\tau)}}{\sum_{k=1}^N \exp{(C_k^\beta s(f_i^v, f_k^t)/\tau)}}, \quad q_{i, j} = \frac{y_{i, j}}{\sum_{k=1}^N y_{i, k}}, \\
    &\quad\mathcal{L}_{sdm}^{i2t} = \text{KL}(\mathbf{p}_i\lVert\mathbf{q}_i) = \frac{1}{N} \sum_{i=1}^N\sum_{j=1}^N p_{i, j}\log\frac{p_{i,j}}{q_{i,j} + \epsilon},
\end{split}
\end{equation}
where $f_i^v$ and $f_j^t$ are the representations of the $i$-th image and the $j$-th text, respectively, $y_{i, j}$ is the ground truth of matching label,  $y_{i, j}=1$ means that $(f_i^v, f_j^t)$ is a matched pair from the same identity, while $y_{i, j}=0$ indicates the unmatched pair, $s(f_i^v, f_k^t)$ measures the cosine similarity between $f_i^v$ and $f_j^t$, $\tau$ is a learned temperature, and $\epsilon$ is a small number to avoid numerical problems.
Symmetrically, we apply a similar approach to incorporate the confidence score into the SDM loss for image-to-text matching $\mathcal{L}_{sdm}^{t2i}$.
The overall SDM loss can be formulated as:
\begin{equation}
\mathcal{L}_{sdm} = \mathcal{L}_{sdm}^{i2t} + \mathcal{L}_{sdm}^{t2i}.
\end{equation}

Finally, ID loss~\cite{zheng2020dual} encourages the model to categorize the visual representations and the textual ones of the same identity into the same class, so as to enhance the feature compactness of each class.
Since the noise stems from the generated pseudo texts, we only modify the loss for text classification by incorporating the confidence score, while keeping the loss for image classification $\mathcal{L}_{id}^{i}$ unchanged.
The confidence score-based ID loss for text classification is defined as:
\begin{equation}
\mathcal{L}_{id}^{t} = -\frac{1}{MN} 
\sum_{i=1}^N\sum_{j=1}^M y_j\log\frac{\exp{(C_i^\beta \mathbf{W}_{j}f_i^t + b_{j}})}{\sum_{k=1}^M \exp{(C_i^\beta \mathbf{W}_{k}f_i^t + b_k})},
\end{equation}
where $y$ is a one-hot label, in which the ground-truth class from the total $M$ classes has a probability of $1$, $\mathbf{W}$ and $b$ are the weight matrix and bias of the classification head, respectively. 
The overall ID loss is denoted as:
\begin{equation}
\mathcal{L}_{id} = \mathcal{L}_{id}^{i} + \mathcal{L}_{id}^{t}.
\end{equation}

\subsection{Visualization}
\label{visualization}
Figure~\ref{fig5} shows the comparison of the generated results from object detection by BUTD~\cite{anderson2018bottom}, image captioning on the vision-language model BLIP~\cite{li2022blip} and the proposed FineIC strategy.
It is clear from the figure that the results from object detection are inferior due to the lack of fine-grained detection.
And image captioning only exhibits a superficial depiction, which is also a sub-optimal solution for this particular fine-grained task.
In contrast, the proposed FineIC presents significantly enriched descriptions with abundant fine-grained attributes of the person images.
Meanwhile, compared to the rigid generated texts using the hand-crafted template in $\mu$-TBPS$^+$, the descriptions from the finetuned language model in $\mu$-TBPS exhibit a more diverse and flexible superiority.
The visualization results vividly demonstrate the effectiveness of the proposed FineIC strategy in GTR.

\end{sloppypar}
\end{document}